%% file: main.tex
\RequirePackage[svgnames]{xcolor}

\documentclass[11pt,letterpaper]{mystyle}

\usepackage[all]{hypcap}
\usepackage[svgnames]{xcolor}
\usepackage[comma,authoryear,compress]{natbib}
\usepackage{hyperref}[citecolor=lightblue]

\hypersetup{
    colorlinks = true,
    citecolor = {YaleBlue},
}

\usepackage{algorithm}
\usepackage{algorithmicx}
\usepackage{algpseudocode}
\usepackage{microtype}
\usepackage{graphicx}
\expandafter\def\csname ver@subfig.sty\endcsname{}
\usepackage{booktabs}
\usepackage{float}
\usepackage{bigstrut}

\usepackage{amsmath}
\usepackage{amssymb}
\usepackage{mathtools}
\usepackage{amsthm}
\usepackage{mathrsfs}
\usepackage{nicefrac}
\usepackage{dsfont}
\usepackage{enumitem}
\usepackage{subcaption}
\usepackage{graphicx,subfig}
\usepackage{cleveref}
\usepackage{bxcoloremoji}

\usepackage{float}

\usepackage[utf8]{inputenc}
\usepackage[T1]{fontenc}
\usepackage{hyperref}
\usepackage{url}
\usepackage{booktabs}
\usepackage{amsfonts}
\usepackage{nicefrac}
\usepackage{microtype}
\usepackage{graphicx}
\usepackage{subcaption}
\usepackage{amssymb}
\usepackage{fdsymbol}
\usepackage{wrapfig}
\usepackage{lipsum}
\usepackage{enumitem}
\usepackage{stackengine}
\usepackage[font=small,labelfont=bf]{caption}
\usepackage{color}
\usepackage{adjustbox}

\usepackage{rotating}
\usepackage{makecell}

\input{macro}

\newtcolorbox{AIbox}[2][]{aibox,title=#2,#1}
\definecolor{lightblue}{rgb}{0.22,0.45,0.70}%
\definecolor{Gray}{gray}{0.95}
\definecolor{Cornsilk}{rgb}{1.0, 0.97, 0.86}

\usepackage{amsmath}

\usepackage[all]{hypcap}

\title{DualCoT-VLA: Visual-Linguistic Chain of Thought via Parallel Reasoning for Vision-Language-Action Models}

\runningtitle{DualCoT-VLA: Visual-Linguistic Chain of Thought via Parallel Reasoning for VLA Models}

\makeatletter
\renewcommand\AB@affilsepx{\\\protect\Affilfont}
\makeatother
\author[1]{Zhide Zhong$^*$}
\author[1]{Junfeng Li$^*$}
\author[1]{Junjie He$^*$}
\author[1]{Haodong Yan}
\author[1]{Xin Gong}
\author[2]{Guanyi Zhao}
\author[2]{Yingjie Cai}
\author[2]{Jiantao Gao}
\author[2]{Xu Yan}
\author[2]{Bingbing Liu}
\author[1]{Yingcong Chen}
\author[1]{Liuqing Yang}
\author[1]{Haoang Li$^\dagger$}
\affil[1]{The Hong Kong University of Science and Technology (Guangzhou)}
\affil[2]{Huawei Foundation Model Department}

\projectpage{\textbf{Project Page:} \url{https://livfour.github.io/DualCoT-VLA/}}

\begin{document}

\input{sections/abstract}

\maketitle
\vspace{3mm}
\input{sections/introduction}
\input{sections/relatedwork}
\input{sections/method}
\input{sections/experiments}
\input{sections/discussions}
\input{sections/conclusion}
\input{sections/acknowledge}
\clearpage
\bibliography{main}

\appendix
\input{sections/appendix}
\appendix
\end{document}

%% file: macro.tex
\newcommand{\x}{\mathbf{x}}

\definecolor{blanchedalmond}{rgb}{1.0, 0.92, 0.8}
\definecolor{carmine}{rgb}{0.59, 0.0, 0.09}
\definecolor{lightblue}{rgb}{0.22,0.45,0.70}

\renewcommand{\mathbf}{\boldsymbol}

\makeatletter
\def\Ddots{\mathinner{\mkern1mu\raise\p@
\vbox{\kern7\p@\hbox{.}}\mkern2mu
\raise4\p@\hbox{.}\mkern2mu\raise7\p@\hbox{.}\mkern1mu}}
\makeatother

\definecolor{amaranth}{rgb}{0.9, 0.17, 0.31}
\definecolor{antiquebrass}{rgb}{0.8, 0.58, 0.46}
\definecolor{antiquefuchsia}{rgb}{0.57, 0.36, 0.51}
\definecolor{chromeyellow}{rgb}{0.31, 0.47, 0.26}

\usepackage{xspace}
\newcommand{\ourmethod}{DualCoT-VLA\xspace}

%% file: sections/abstract.tex
\begin{abstract}
Vision-Language-Action (VLA) models map visual observations and language instructions directly to robotic actions. While effective for simple tasks, standard VLA models often struggle with complex, multi-step tasks requiring logical planning, as well as precise manipulations demanding fine-grained spatial perception. Recent efforts have incorporated Chain-of-Thought (CoT) reasoning to endow VLA models with a ``thinking before acting'' capability. However, current CoT-based VLA models face two critical limitations: 1) an inability to simultaneously capture low-level visual details and high-level logical planning due to their reliance on isolated, single-modal CoT; 2) high inference latency with compounding errors caused by step-by-step autoregressive decoding. To address these limitations, we propose \ourmethod, a visual-linguistic CoT method for VLA models with a parallel reasoning mechanism. To achieve comprehensive multi-modal reasoning, our method integrates a visual CoT for low-level spatial understanding and a linguistic CoT for high-level task planning. Furthermore, to overcome the latency bottleneck, we introduce a parallel CoT mechanism that incorporates two sets of learnable query tokens, shifting autoregressive reasoning to single-step forward reasoning. Extensive experiments demonstrate that our \ourmethod achieves state-of-the-art performance on the LIBERO and RoboCasa GR1 benchmarks, as well as in real-world platforms.
\end{abstract}

%% file: sections/introduction.tex
\section{Introduction}
\label{sec:intro}

Vision-Language-Action (VLA)~\citep{BlackK-RSS-25,black2025pi,kim2024openvla,KimM1-RSS-25,song2025reconvla} models currently represent a leading approach for robotic manipulation. However, these models fundamentally operate as direct mappings from vision and language to actions. Consequently, they often struggle with complex tasks that require long-horizon logical planning and precise spatial perception. To mitigate this, recent work has incorporated Chain-of-Thought (CoT) reasoning into VLA models~\citep{zhao2025cot,huangthinkact,lv2025f1,zhangdreamvla,yin2025deepthinkvla}, demonstrating improved task generalization.

Despite these improvements, current CoT-based VLAs~\citep{zhao2025cot,huangthinkact,lv2025f1,zhangdreamvla,yin2025deepthinkvla,huang2026fast} face two major limitations. First, existing methods typically confine their intermediate CoT traces to a single modality, failing to simultaneously achieve robust long-horizon logical planning and spatial perception. Specifically, linguistic-only CoT offers logical foresight but lacks precise spatial perception, whereas visual CoT captures spatial details but struggles with long-horizon logical planning. Second, most of them rely on the autoregressive (AR) inference of discrete tokens. This step-by-step inference introduces severe latency and cumulative errors. 

To address the above limitations, we propose \textbf{\ourmethod}, a visual-linguistic CoT paradigm that reasons in the continuous latent space. Within this paradigm, a visual CoT extracts low-level 3D spatial cues, while a linguistic CoT formulates high-level logical plans for comprehensive multimodal reasoning. To overcome inference latency, we further introduce a parallelized CoT mechanism.

\begin{figure*}[!t]
\centering
\includegraphics[width=1.0\linewidth]{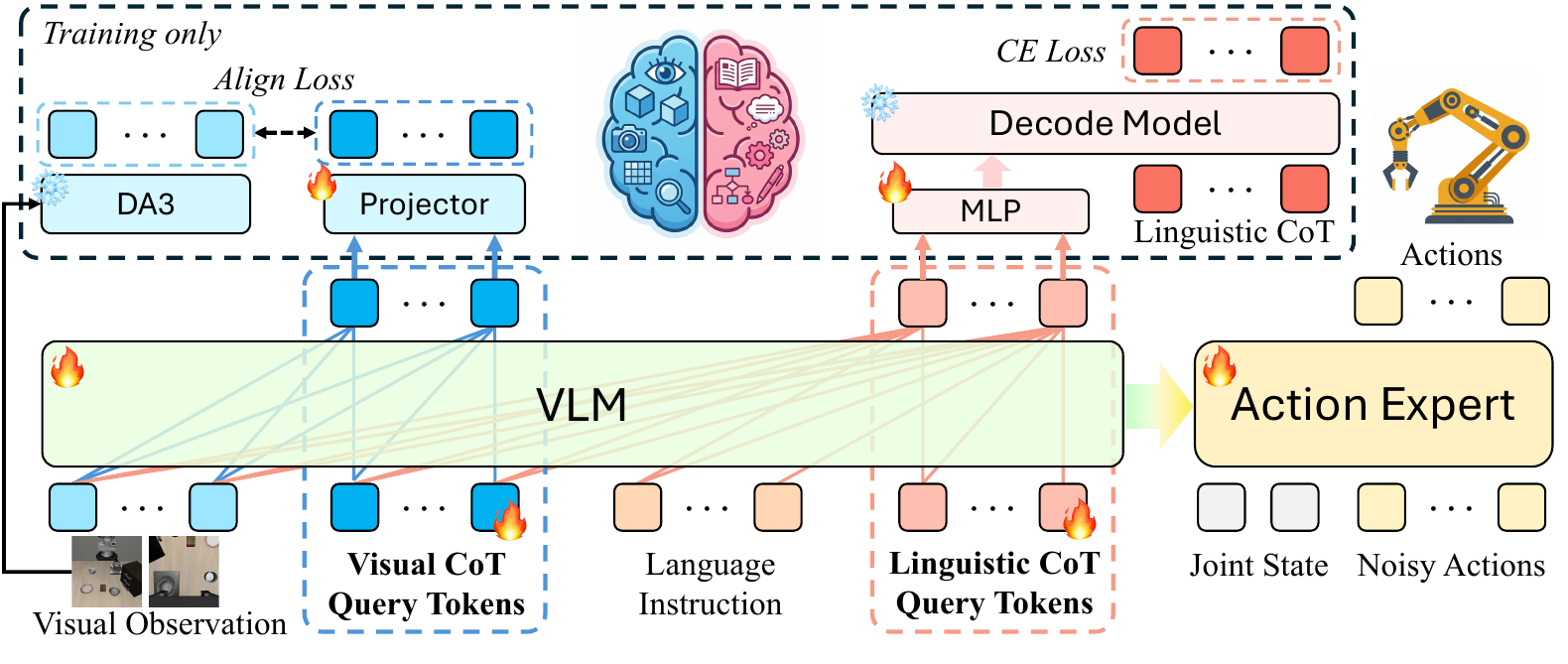}
\caption{\textbf{Overview of our \ourmethod.} The VLM backbone processes a unified sequence comprising visual observations, language instructions, and two sets of learnable query tokens. our \ourmethod employs a visual-linguistic CoT for VLA models: (1) Visual CoT: visual query hidden states are aligned with dense spatial features from a frozen Depth Anything 3~\citep{lin2025depth}; (2) Linguistic CoT: linguistic query hidden states act as conditioning prefixes for a frozen auxiliary LLM to predict explicit CoT text. Finally, the  reasoning-enriched hidden states guide a Flow-Matching DiT (Action Expert) to predict continuous action. During inference, the auxiliary modules are discarded for highly efficient, single-forward-pass action generation.}
\label{fig:pipeline}
\end{figure*}

Specifically, as shown in Fig.~\ref{fig:pipeline}, our \ourmethod employs a \textbf{parallel} reasoning mechanism, avoiding the significant inference latency and compounding errors inherent in autoregressive CoT. Furthermore, we adopt \textbf{implicit} CoT rather than forcing the model to decode explicit reasoning content, as explicit CoT inherently suffers from severe information redundancy and requires time-consuming multi-step decoding~\citep{hao2024training, wei2025sim}. We implement the implicit parallel CoT reasoning by incorporating two sets of learnable CoT query tokens that enable the Vision-Language Model (VLM)~\citep{liu2023llava} backbone of the VLA model to drive condensed reasoning-aware representations in a single forward pass. To perform comprehensive multimodal CoT reasoning, we propose a \textbf{visual-linguistic CoT} paradigm for VLA models that leverages auxiliary models to supervise the VLM backbone of VLA models to acquire implicit reasoning capabilities. To endow the VLM with visual reasoning capability, we distill visual priors from a frozen Vision Foundation Model (VFM)~\citep{simeoni2025dinov3, lin2025depth} by aligning the output representations corresponding to the visual CoT query tokens with the VFM-encoded visual features. Concurrently, to facilitate linguistic reasoning capability, we align the output representations corresponding to linguistic CoT query tokens with the representation space of a frozen auxiliary Large Language Model (LLM)~\citep{brown2020language, bai2025qwen3}. By tasking the auxiliary LLM to generate complete CoT text conditioned on the output representations corresponding to linguistic CoT query tokens, we force the VLM backbone to internalize compressed logical planning within the continuous latent space. Finally, these reasoning-enriched hidden states are fed into a downstream action expert for action prediction. During inference, the auxiliary teacher modules are discarded, allowing the VLM to think comprehensively in a single forward pass and drive the action expert at a high frequency.

We comprehensively evaluate \ourmethod across both simulation benchmarks and real-world platforms. Experiments on the LIBERO~\citep{liu2023libero} and RoboCasa GR1~\citep{bjorck2025gr00t} benchmarks demonstrate that our method achieves state-of-the-art (SOTA) performance. Furthermore, real-world robot experiments confirm that our model seamlessly transfers its robust task planning and 3D spatial perception to real-world  environments. Our work makes the following contributions:
\begin{itemize}
\item We propose a Visual-Linguistic Chain-of-Thought paradigm, which endows the VLA model with both low-level spatial perception and high-level logical planning capabilities.

\item We introduce \ourmethod, an efficient instantiation of this paradigm that incorporates a parallel CoT mechanism. This mechanism bypasses the slow, autoregressive CoT decoding, enabling single-step inference that accelerates downstream action prediction.

\item Extensive evaluations demonstrate that \ourmethod achieves state-of-the-art performance on the challenging LIBERO and RoboCasa GR1 benchmarks, as well as in real-world deployments.

\end{itemize}

%% file: sections/relatedwork.tex
\section{Related Work}
\label{sec:related}

\subsection{Single-Modal Chain-of-Thought in VLA models}
To enhance standard VLA models, researchers have incorporated Chain-of-Thought (CoT), enabling robots to ``think before acting''~\citep{zawalski2025robotic}. However, current approaches generally decouple reasoning into single modalities. Linguistic CoT prompts VLMs to generate textual sub-goals or reasoning traces (e.g., Embodied CoT~\citep{zawalski2025robotic}, ThinkAct~\citep{huangthinkact}). Although effective for high-level logic, these methods often lack physical grounding, as the text is frequently ambiguous regarding precise spatial coordinates and object poses. Conversely, Visual CoT forces models to explicitly forecast sub-goal images or dense optical flows (e.g., CoT-VLA~\citep{zhao2025cot}, FlowVLA~\citep{zhong2025flowvla}). Although capturing visual details, these methods struggle with abstract planning and are computationally expensive. Consequently, existing methods fail to synergize high-level semantic logic with low-level spatial perception. A concurrent work~\citep{bai2026latent}, LaRA-VLA (Bai et al., 2026), also explores multi-modal reasoning. Unlike LaRA-VLA, our method maintains supervision on the implicit textual CoT, avoiding collapse and retaining the ability to decode these latent tokens into explicit text during inference. Built upon this design, our proposed DualCoT-VLA bridges the aforementioned gap by unifying both modalities, enabling the robot to perform comprehensive Visual-Linguistic reasoning.

\subsection{Autoregressive Reasoning in VLA models}
Beyond modality, the sequential nature of current reasoning mechanisms presents a critical bottleneck. Most CoT approaches rely heavily on autoregressive decoding, where reasoning steps are generated token-by-token. 
This limitation is most severe in explicit methods~\citep{zhao2025cot}, where the sequential generation of multi-modal discrete tokens incurs severe inference latency. Furthermore, explicitly generating lengthy reasoning traces makes the system highly susceptible to cumulative errors; a single hallucinated text token or distorted visual prediction can cascade and derail the entire execution.
To mitigate these issues, recent work has explored implicit reasoning strategies. Approaches like Coconut~\citep{hao2024training} in NLP and Fast-ThinkAct~\citep{huang2026fast} in robotics compress thoughts into continuous latent states to avoid the overhead of explicit token generation. 
However, despite removing explicit output, these methods typically retain an underlying autoregressive structure in the latent space, where the generation of the current thought vector still depends on the previous one. This sequential dependency fundamentally limits the speed of inference and prevents full parallelization.
Our \ourmethod breaks this autoregressive constraint by employing parallelized CoT query tokens. We simultaneously distill robust 3D spatial priors and internalize high-level linguistic plans in a single forward pass. This design eliminates the risk of token cascading errors and significantly reduces inference latency.

%% file: sections/method.tex
\section{Methodology}
\label{sec:method}

In this section, we detail the proposed \ourmethod framework. We first provide an overview of the architecture (Sec.~\ref{subsec:overview}). Next, we present the parallel implicit CoT mechanism based on learnable query tokens (Sec.~\ref{subsec:sequence}). We then delve into the two core reasoning streams: the implicit Visual CoT via geometric distillation (Sec.~\ref{subsec:visual_cot}) and the implicit Linguistic CoT via step-level supervision (Sec.~\ref{subsec:linguistic_cot}). Finally, we introduce the action head and final loss function (Sec.~\ref{subsec:action_generation}).

\subsection{Overview}
\label{subsec:overview}
As shown in Fig.~\ref{fig:pipeline}, our \ourmethod consists of three core components: a Vision-Language Model (VLM) backbone that processes multimodal inputs, a dual-stream implicit CoT mechanism (comprising visual and linguistic paths), and a downstream Diffusion Transformer (DiT)~\citep{peebles2023scalable} action head. During the training stage, for the Visual CoT, the hidden states derived from the visual query tokens are aligned with the dense features of Depth Anything 3 (DA3)~\citep{lin2025depth} encoder, enabling the VLM backbone to reason about low-level spatial information from the observed image. For the Linguistic CoT, the hidden states corresponding to the linguistic query tokens are utilized as conditioning prefixes for an auxiliary LLM, equipping the VLM backbone with the capability to efficiently perform logical task planning. During the inference stage, these auxiliary modules are discarded, allowing the model to bypass slow autoregressive decoding and predict CoT-informed actions in a single forward pass.

\subsection{Parallel Implicit CoT via Learnable Query Tokens}
\label{subsec:sequence}
We use learnable query tokens to guide the VLM backbone in extracting and reasoning over effective representations from both visual and linguistic modalities. Given an image $\mathbf{I}$ and a language instruction $\mathbf{L}$, we construct a unified sequence to serve as VLM backbone input. 

We introduce two sets of distinct learnable query tokens: a set of visual CoT query tokens $\mathbf{Q}_{\textnormal{vis}} \in \mathbb{R}^{M \times d_{\textnormal{VLM}}}$ (where $M=16$), and a set of linguistic CoT query tokens $\mathbf{Q}_{\textnormal{lin}} \in \mathbb{R}^{N \times d_{\textnormal{VLM}}}$ (where $N=4$), with $d_{\textnormal{VLM}}$ denoting the hidden dimension of the VLM backbone. We formulate the unified input sequence $\mathbf{X}_{\textnormal{input}}$ by interleaving the encoded multimodal inputs with these query sequences in the following order:
\begin{equation}
    \mathbf{X}_{\textnormal{input}} = \Big[ \mathbf{V}_{\textnormal{obs}} , \mathbf{Q}_{\textnormal{vis}} , \mathbf{L}_{\textnormal{instr}} , \mathbf{Q}_{\textnormal{lin}} \Big]
\end{equation}
where $\mathbf{V}_{\textnormal{obs}}$ denotes the encoded visual observation tokens derived from $\mathbf{I}$, and $\mathbf{L}_{\textnormal{instr}}$ represents the language instruction tokens tokenized from $\mathbf{L}$. 

By passing $\mathbf{X}_{\textnormal{input}}$ through the VLM backbone, the query sequences interact seamlessly with the rich multimodal context via self-attention mechanisms. The backbone simultaneously outputs the corresponding hidden states for both reasoning streams in a single forward pass, denoted as $\mathbf{H}_{\textnormal{vis}} \in \mathbb{R}^{M \times d_{\textnormal{VLM}}}$ and $\mathbf{H}_{\textnormal{lin}} \in \mathbb{R}^{N \times d_{\textnormal{VLM}}}$, which encapsulate the implicit visual and linguistic CoT features, respectively.

\begin{figure*}[t]
\centering
\sbox0{\includegraphics[width=0.55\textwidth]{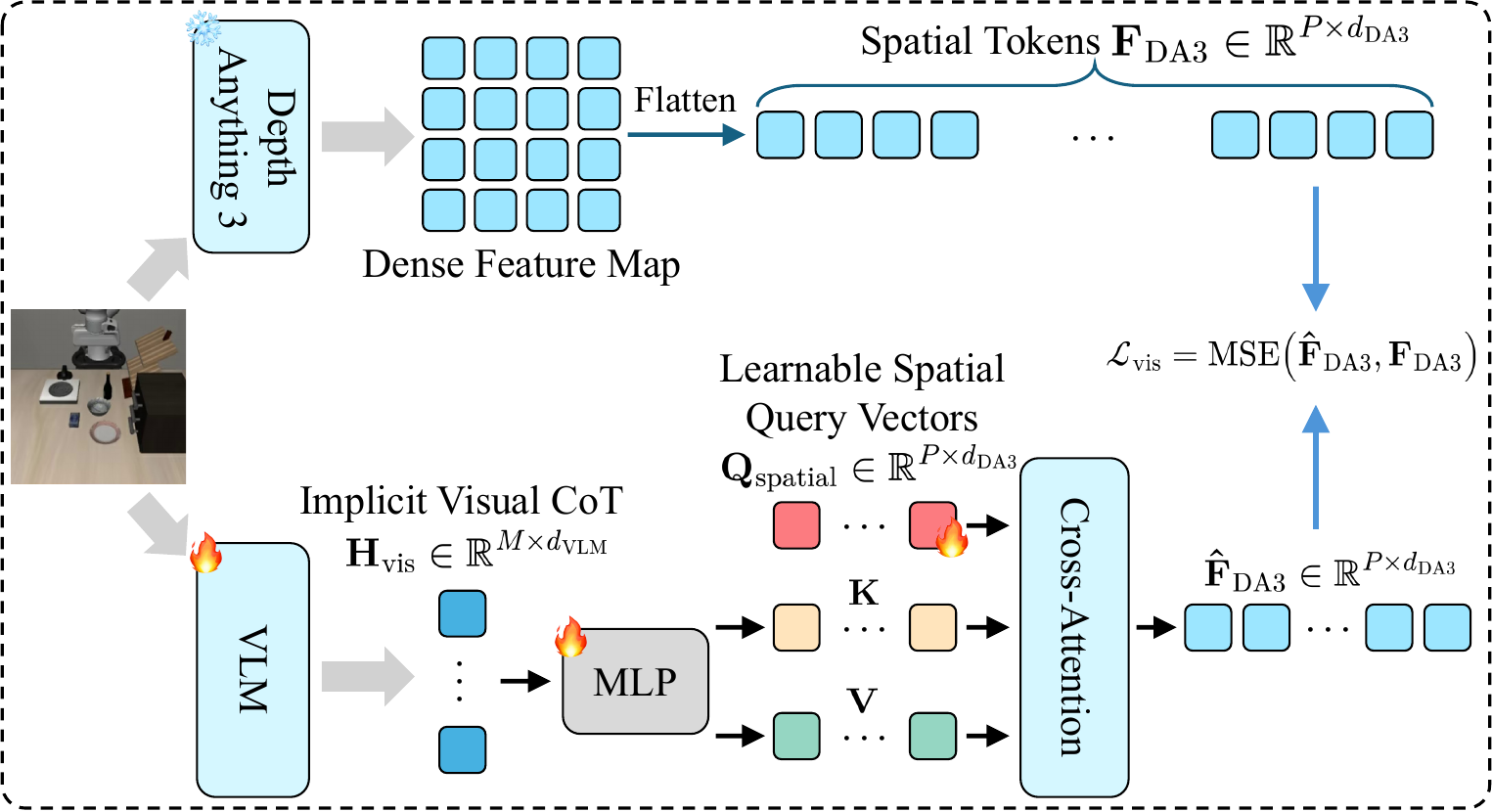}}%
\begin{subfigure}[t]{0.55\textwidth}
\centering
\usebox0
\caption{Illustration of the Visual CoT. Learnable spatial query vectors decode compressed output hidden states of visual CoT tokens into a high-resolution feature map via cross-attention, supervised by the dense features from a frozen DA3 using an MSE loss.}
\label{fig:visual_cot}
\end{subfigure}
\hfill
\begin{subfigure}[t]{0.43\textwidth}
\centering
\includegraphics[height=\ht0]{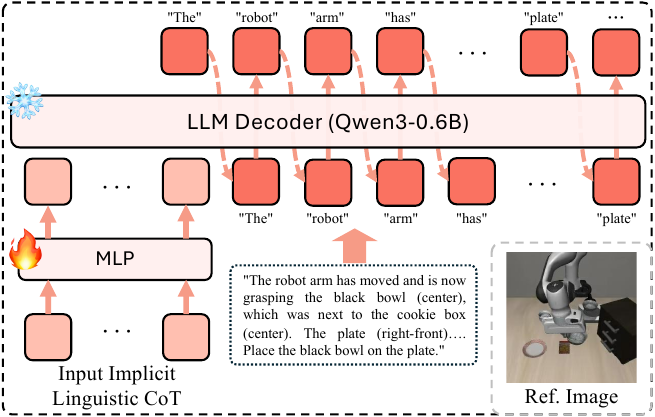}
\caption{Illustration of the Linguistic CoT. A frozen auxiliary LLM decodes continuous hidden states of linguistic CoT query tokens into explicit task planning sequences.}
\label{fig:linguistic_cot}
\end{subfigure}
\caption{\textbf{Detailed illustrations of the dual-stream implicit CoT mechanism.} (a) Visual CoT via geometric distillation from a frozen Depth Anything 3 encoder. (b) Linguistic CoT via step-level supervision from a frozen auxiliary LLM.}
\label{fig:cot_detail}
\end{figure*}

\subsection{Visual CoT}
\label{subsec:visual_cot}
As shown in Fig.~\ref{fig:cot_detail}(a), during training, we use a frozen DA3 model as the geometric teacher. Given the observed image $\mathbf{I}$, the DA3 model extracts a dense visual feature map, which is subsequently flattened into a sequence of spatial tokens $\mathbf{F}_{\textnormal{DA3}} \in \mathbb{R}^{P \times d_{\textnormal{DA3}}}$, where $P$ denotes the number of spatial patches. A naive point-wise alignment between the highly compressed visual CoT tokens $\mathbf{H}_{\textnormal{vis}}$ (where $M=16$) and the dense feature map $\mathbf{F}_{\textnormal{DA3}}$ is architecturally infeasible due to the dimension mismatch. To address this mismatch, we used a cross-attention-based projector.  

Specifically, we initialize a set of learnable spatial query vectors, $\mathbf{Q}_{\textnormal{spatial}} \in \mathbb{R}^{P \times d_{\textnormal{DA3}}}$, that match the spatial resolution of the teacher features. Within the projector, these queries act as information probes, cross-attending to the visual CoT tokens $\mathbf{H}_{\textnormal{vis}}$ to reconstruct the dense representations:
\begin{equation}
    \mathbf{\hat{F}}_{\textnormal{DA3}} = \textnormal{CrossAttention}\big( \mathbf{Q}_{\textnormal{spatial}}, \mathcal{P}(\mathbf{H}_{\textnormal{vis}}), \mathcal{P}(\mathbf{H}_{\textnormal{vis}}) \big)
\end{equation}
where $\mathcal{P}$ are linear projections aligning the VLM hidden dimension with the teacher dimension. The geometric distillation is then optimized using a Mean Squared Error (MSE) loss:
\begin{equation}
    \mathcal{L}_{\textnormal{vis}} = \text{MSE}\big( \mathbf{\hat{F}}_{\textnormal{DA3}}, \mathbf{F}_{\textnormal{DA3}} \big)
\end{equation}
This geometric distillation process compels the VLM backbone to encode low-level spatial priors into a highly compact latent representation (16 continuous hidden states). 

\subsection{Linguistic CoT}
\label{subsec:linguistic_cot}
As illustrated in Fig.~\ref{fig:cot_detail}(b), we supervise the linguistic CoT stream by using a frozen auxiliary language model to decode the backbone's hidden states $\mathbf{H}_{\textnormal{lin}}$ into human-readable task planning sequences $\mathbf{Y}_{\textnormal{cot}} = (y_1, y_2, \dots, y_L)$ of length $L$. Specifically, we adopt a lightweight language model (e.g., Qwen3-0.6B~\citep{bai2025qwen3}) as this frozen text decoder. Because the main VLM backbone typically operates at a higher dimensionality, we first apply a learnable linear projector $\mathcal{P}_{\textnormal{lin}}$ to align the dimensions of $\mathbf{H}_{\textnormal{lin}}$ with the auxiliary decoder. 

During training, the projected states $\mathcal{P}_{\textnormal{lin}}(\mathbf{H}_{\textnormal{lin}})$ are prepended as prefix tokens to the embedded CoT text sequence. Thanks to the causal self-attention mechanism of the autoregressive decoder, these prefix states condition the generation of the explicit CoT text. We optimize the VLM backbone and the projector $\mathcal{P}_{\textnormal{lin}}$ using the standard Cross-Entropy (CE) loss:
\begin{equation}
    \mathcal{L}_{\textnormal{lin}} = - \sum_{i=1}^{L} \log p_{\phi} \big( y_i \mid \mathcal{P}_{\textnormal{lin}}(\mathbf{H}_{\textnormal{lin}}), y_{<i} \big)
\end{equation}
where $\phi$ denotes the parameters of the frozen auxiliary decoder. By enforcing this text reconstruction task, we provide step-level implicit supervision, compelling the VLM backbone to deeply internalize logical task planning into its $N=4$ compact continuous tokens, explicitly avoiding latent representation collapse.

\subsection{Flow-Matching Action Head}
\label{subsec:action_generation}
To predict continuous action chunks, we employ a Diffusion Transformer (DiT) action head trained via a Flow Matching objective. First, the VLM backbone processes the entire multimodal prompt, including visual observations, language instruction, and visual and linguistic CoT query tokens. We extract the full sequence of hidden states from the \textbf{final layer} of the VLM, denoted $\mathbf{H}_{\textnormal{vlm}}$, which serves as a unified conditioning context that incorporates both spatial grounding and logical planning.

During the training stage, let $\mathbf{A}$ represent the ground-truth action chunk. Following standard Flow Matching~\citep{lipman2022flow} formulations, we train the DiT, parameterized by $\theta$, to predict a vector field $v_\theta$ that transports a standard Gaussian noise $\mathbf{a}_0 \sim \mathcal{N}(\mathbf{0}, \mathbf{I})$ to the target action distribution. The DiT takes the current noisy action $\mathbf{a}_t$, the continuous time step $t \sim \mathcal{U}(0, 1)$, and the robot's state as its primary sequence inputs. Moreover, the reasoning context $\mathbf{H}_{\textnormal{vlm}}$ is injected into the DiT blocks via cross-attention, where the action expert queries the VLM's hidden states of finaly layer for geometric and logical guidance. The action predication is supervised using a MSE loss to match the target vector field $(\mathbf{A} - \mathbf{a}_0)$:
\begin{equation}
    \mathcal{L}_{\textnormal{act}} = \mathbb{E}_{t, \mathbf{a}_0, \mathbf{A}} \Big[ \big\| v_\theta(\mathbf{a}_t, t, \mathbf{H}_{\textnormal{vlm}}) - (\mathbf{A} - \mathbf{a}_0) \big\|^2_2 \Big]
\end{equation}
where $\mathbf{a}_t = t \mathbf{A} + (1 - t) \mathbf{a}_0$ is the linearly interpolated noisy action. 

Finally, the entire framework is optimized end-to-end using a joint loss that balances the visual-linguistic reasoning and the action prediction:
\begin{equation}
    \mathcal{L}_{\textnormal{total}} = \lambda_{\textnormal{vis}} \mathcal{L}_{\textnormal{vis}} + \lambda_{\textnormal{lin}} \mathcal{L}_{\textnormal{lin}} + \lambda_{\textnormal{act}} \mathcal{L}_{\textnormal{act}}
\end{equation}
where $\lambda_{\textnormal{vis}}$, $ \lambda_{\textnormal{lin}}$, and $\lambda_{\textnormal{act}}$ are empirically determined weighting coefficients.

%% file: sections/experiments.tex
\section{Experiments}
\label{sec:experiments}

\begin{figure*}[!t]
\centering
\includegraphics[width=1.0\linewidth]{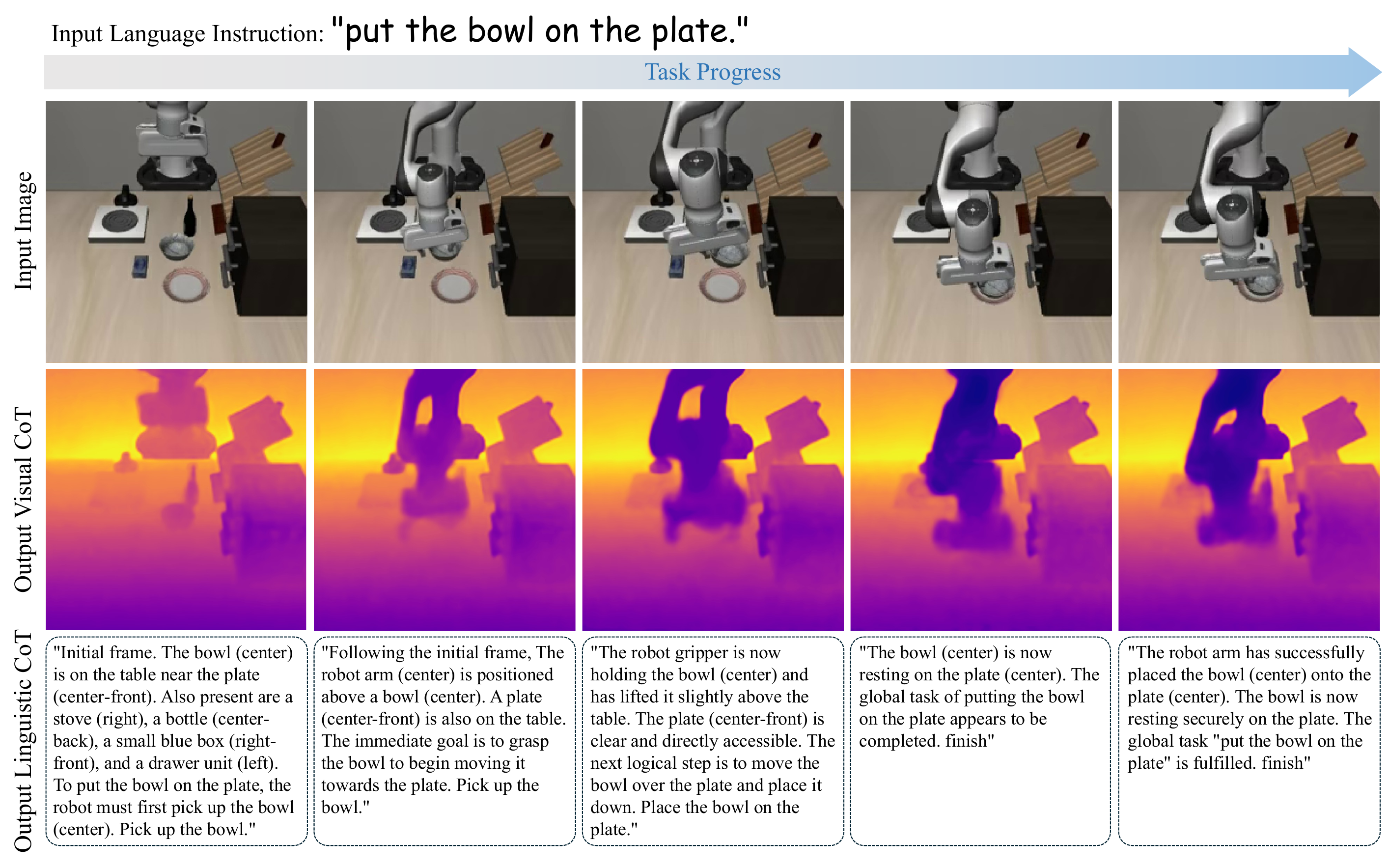}
\caption{\textbf{Qualitative result of our Visual-Linguistic CoT on LIBERO benchmark~\citep{liu2023libero}.} During a robotic manipulation task, given the \textbf{Language Instruction:} ``placing the bowl on the plate'' and \textbf{Input Image}, we visualize the \textbf{Ouput Visual CoT} and \textbf{Output Linguisitic CoT}. To visualize the implicit visual reasoning, we employ a lightweight visual probe to project the compressed hidden states of the visual query tokens back into depth maps. This visualization demonstrates that the continuous latent tokens successfully capture dense 3D geometric structures and spatial priors. The explicitly decoded Linguistic CoT text, demonstrating the model's capacity for scene comprehension, dynamic state tracking, and step-level logical planning.}
\label{fig:exp_vis}
\end{figure*}

In this section, we comprehensively evaluate the proposed \ourmethod framework to answer the following four key questions: 
\begin{itemize}
    \item \textbf{Q1:} Does \ourmethod achieve state-of-the-art (SOTA) performance across diverse and complex robotic manipulation benchmarks and real-world environments?
    \item \textbf{Q2:} Does the implicit reasoning mechanism successfully endow the VLA model with comprehensible visual and linguistic reasoning capabilities?
    \item \textbf{Q3:} What are the individual contributions of the Visual CoT and Linguistic CoT?
    \item \textbf{Q4:} How does the single-step parallel CoT improve inference speed compared to autoregressive CoT models?
\end{itemize}

\subsection{Experimental Setup}
\label{subsec:exp_setup}

\noindent\textbf{Benchmarks.}
To comprehensively evaluate our method, we select two challenging simulation benchmarks with distinct embodiments, alongside real-world physical robot deployments.
\begin{itemize}
    \item \textbf{LIBERO}~\citep{liu2023libero}: A comprehensive benchmark features a 7-DoF robotic arm performing four task suites: \texttt{Spatial}, \texttt{Object}, \texttt{Goal}, and \texttt{Long}.
    \item \textbf{RoboCasa GR1 Tabletop Tasks}~\citep{bjorck2025gr00t}: A highly complex benchmark requiring the control of a 29-DoF dexterous humanoid hand. This environment provides a test for a model's fine-grained spatial perception and high-dimensional action coordination.
    \item \textbf{Real-world Experiments based on AgileX Cobot}: We conduct real-world experiments using an AgileX Cobot dual-arm robot based on the Mobile ALOHA system design~\citep{fu2024mobile}. The robot features 7-DoF arms equipped with parallel grippers, relying entirely on onboard RGB cameras (front-facing and wrist-mounted) for visual input. 
    We design three tabletop manipulation tasks of increasing complexity to rigorously test the model's spatial reasoning and long-horizon multi-step execution capabilities: (1) \textbf{Easy}: picking up a piece of bread and placing it onto a plate; (2) \textbf{Medium}: sequentially grasping two distinct blocks and placing them onto a plate; and (3) \textbf{Hard}: gathering three different fruits and placing them into a target receptacle. These tasks are categorized based on their horizon length.
    For each task, we collect 100 human-teleoperated demonstrations for fine-tuning, effectively evaluating the model's adaptability to unstructured physical environments.
\end{itemize}

\noindent\textbf{Baselines.}
We compare our \ourmethod against a diverse SOTA methods, which can be broadly categorized into: 

\begin{itemize}
    \item \textbf{Non-CoT VLA Models}: Diffusion Policy~\citep{chi2025diffusion}, OpenVLA~\citep{kim2024openvla}, $\pi_0$~\citep{black2025pi}, PD-VLA~\citep{song2025pd}, and the GR00T series~\citep{bjorck2025gr00t}), which directly map observations to actions without intermediate reasoning.
    \item \textbf{Autoregressive (AR) CoT VLA Models}: CoT-VLA~\citep{zhao2025cot}, ThinkAct~\citep{huangthinkact}), DeepThinkVLA~\citep{yin2025deepthinkvla}, Fast-ThinkAct~\citep{huang2026fast}, and LaRA-VLA~\citep{bai2026latent} which autoregressively decode thinking tokens.
\end{itemize}

\noindent\textbf{Implementation Details.}
All training is conducted on NVIDIA H100 GPUs. For model architecture, our \ourmethod utilizes Qwen3-VL-4B~\citep{bai2025qwen3} as the VLM backbone. For the auxiliary teacher modules used during training, we employ Depth Anything 3~\citep{lin2025depth} as the visual teacher model, and the lightweight Qwen3-0.6B~\citep{bai2025qwen3} as the linguistic text decoder. For downstream action generation, we adopt a DiT head parameterized via a Flow Matching paradigm. In terms of the loss weight, we set $\lambda_{vis}$ to 0.1,  $\lambda_{lin}$ to 0.1, and $\lambda_{act}$ to 1.0.

For task-specific hyperparameters, models on LIBERO are trained with a learning rate of 2.5e-5, an action window of 7 steps, and a global batch size of 48. For the RoboCasa GR1, we use a learning rate of 3e-5, an action window of 15 steps, and a global batch size of 256.

\noindent\textbf{Data Construction for Linguistic CoT.}
To provide step-level logical supervision, we utilize explicit text CoT annotations. For the LIBERO benchmark, linguistic CoT annotations are sourced from \citet{yin2025deepthinkvla}. For the RoboCasa GR1 benchmark, we use Qwen3-VL-32B~\citep{bai2025qwen3} to autonomously generate dense CoT annotations from task trajectories. To ensure structured, logical planning, our target text CoT sequence is rigorously formatted to include three components: state tracking of task progress and the robot's physical state, spatial location detailing the absolute and relative geometric positions of task-relevant objects, and action formulation outlining the immediate next action chunk the robot should execute.

\begin{table}[t]
\footnotesize
\centering
\caption{Evaluation results on the LIBERO benchmark~\citep{liu2023libero}. We train our \ourmethod jointly on four suites. We report the success rate (\%) evaluated over 500 episodes for each task. \textbf{Bold} indicates the best result, and \underline{underline} indicates the second best.}
\label{tab:libero_eval}
\setlength{\tabcolsep}{4pt}

\newcolumntype{Y}{>{\centering\arraybackslash}X}
\begin{tabularx}{\textwidth}{l *{5}{Y}}
\toprule
\textbf{Method} & \texttt{Spatial} & \texttt{Object} & \texttt{Goal} & \texttt{Long} & \textbf{Average} \\
\midrule
Diffusion Policy~\citep{chi2025diffusion} & 78.5 & 87.5 & 73.5 & 64.8 & 76.1 \\
OpenVLA~\citep{kim2024openvla} & 84.7 & 88.4 & 79.2 & 53.7 & 76.5 \\
PD-VLA~\citep{song2025pd} & 95.5 & 96.7 & 94.9 & 91.7 & 94.7 \\
$\pi_0$~\citep{black2025pi} & 98.0 & 96.8 & 94.4 & 88.4 & 94.4 \\
$\pi_0$-Fast~\citep{PertschK-RSS-25} & 96.4 & 96.8 & 88.6 & 60.2 & 85.5 \\
$\pi_{0.5}$~\citep{black2025pi} & \underline{98.8} & 98.2 & \underline{98.0} & 92.4 & 96.9 \\
GR00T-N1~\citep{bjorck2025gr00t} & 94.4 & 97.6 & 93.0 & 90.6 & 93.9 \\
GR00T-N1.6~\citep{bjorck2025gr00t} & 97.7 & 98.5 & 97.5 & 94.4 & 97.0 \\
OpenVLA-OFT~\citep{KimM1-RSS-25} & 97.6 & 98.4 & 97.9 & 94.5 & 97.1 \\
CoT-VLA~\citep{zhao2025cot} & 87.5 & 91.6 & 87.6 & 69.0 & 83.9 \\
ThinkAct~\citep{huangthinkact} & 88.3 & 91.4 & 87.1 & 70.9 & 84.4 \\
DeepThinkVLA~\citep{yin2025deepthinkvla} & 96.6 & \underline{99.0} & 96.4 & 96.2 & 97.0 \\
Fast-ThinkAct~\citep{huang2026fast} & 92.0 & 97.2 & 90.2 & 79.4 & 89.7 \\
LaRA-VLA~\citep{bai2026latent} & 96.4 & 98.6 & \textbf{99.8} & \underline{96.6} & \underline{97.9} \\
\midrule
\textbf{\ourmethod (Our)} & \textbf{99.4} & \textbf{99.8} & 97.8 & \textbf{98.2} & \textbf{98.8} \\
\bottomrule
\end{tabularx}
\end{table}

\begin{table*}[t]
\centering
\footnotesize
\caption{Evaluation Results on RoboCasa GR1 Tabletop Tasks~\citep{bjorck2025gr00t}. We train one model jointly on all 24 tasks, and report mean results over 50 rollouts per task. \textbf{Bold} indicates the best result, and \underline{underline} indicates the second best.}
\label{tab:robocasa_eval}
\setlength{\tabcolsep}{3.9pt}
\resizebox{\textwidth}{!}{%
\begin{tabular}{l *{7}{c}}
\toprule
\textbf{Task} & GR00T-N1.5 & GR00T-N1.6 & Qwen3GR00T & Qwen3PI & Qwen3OFT & Qwen3FAST & \textbf{\ourmethod (Our)} \\
\midrule
BottleToCabinetClose & \underline{54.0} & 51.5 & 46.0 & 26.0 & 30.0 & 38.0 & \textbf{66.0} \\
CanToDrawerClose & 50.0 & 13.0 & \textbf{80.0} & 62.0 & \underline{76.0} & 44.0 & 64.0 \\
CupToDrawerClose & 38.0 & 8.5 & \underline{54.0} & 42.0 & 44.0 & \textbf{56.0} & 46.0 \\
MilkToMicrowaveClose & \textbf{60.0} & 14.0 & 48.0 & 50.0 & 44.0 & 44.0 & \underline{58.0} \\
PotatoToMicrowaveClose & 32.0 & \underline{41.5} & 28.0 & \textbf{42.0} & 32.0 & 14.0 & 30.0 \\
WineToCabinetClose & \underline{38.0} & 16.5 & \textbf{46.0} & 32.0 & 36.0 & 14.0 & \underline{38.0} \\
\addlinespace
CuttingboardToBasket & 38.0 & \textbf{58.0} & 48.0 & 40.0 & 50.0 & \underline{54.0} & 44.0 \\
CuttingboardToCardboardbox & 46.0 & \underline{46.5} & 40.0 & 46.0 & 40.0 & 42.0 & \textbf{54.0} \\
CuttingboardToPan & 58.0 & 68.5 & 68.0 & 60.0 & \underline{70.0} & 58.0 & \textbf{80.0} \\
CuttingboardToPot & 62.0 & \textbf{65.0} & 52.0 & 40.0 & 54.0 & 58.0 & \underline{64.0} \\
CuttingboardToTieredbasket & 28.0 & \underline{46.5} & \textbf{56.0} & 44.0 & 38.0 & 40.0 & 46.0 \\
\addlinespace
PlacematToBasket & 30.0 & \textbf{58.5} & 42.0 & 44.0 & 32.0 & 36.0 & \underline{48.0} \\
PlacematToBowl & \textbf{60.0} & 57.5 & 44.0 & 52.0 & \underline{58.0} & 38.0 & \underline{58.0} \\
PlacematToPlate & 56.0 & \underline{63.0} & 48.0 & 50.0 & 52.0 & 42.0 & \textbf{74.0} \\
PlacematToTieredshelf & \textbf{36.0} & \underline{28.5} & 18.0 & 28.0 & 24.0 & 18.0 & 26.0 \\
\addlinespace
PlateToBowl & 52.0 & \underline{57.0} & \textbf{60.0} & 52.0 & \textbf{60.0} & 52.0 & 50.0 \\
PlateToCardboardbox & 48.0 & 43.5 & \underline{50.0} & 40.0 & \underline{50.0} & 30.0 & \textbf{56.0} \\
PlateToPan & 60.0 & 51.0 & 54.0 & 36.0 & \underline{66.0} & 48.0 & \textbf{70.0} \\
PlateToPlate & 52.0 & \textbf{78.7} & 70.0 & 48.0 & 68.0 & 50.0 & \underline{76.0} \\
\addlinespace
TrayToCardboardbox & 32.0 & \underline{51.5} & 38.0 & 34.0 & 44.0 & 28.0 & \textbf{52.0} \\
TrayToPlate & 58.0 & \textbf{71.0} & 56.0 & \underline{64.0} & 56.0 & 34.0 & \underline{64.0} \\
TrayToPot & 44.0 & \underline{64.5} & 50.0 & 44.0 & 62.0 & 46.0 & \textbf{70.0} \\
TrayToTieredbasket & \textbf{60.0} & \underline{57.0} & 36.0 & 50.0 & 54.0 & 36.0 & \textbf{60.0} \\
TrayToTieredshelf & \textbf{64.0} & \underline{31.5} & 16.0 & 28.0 & 30.0 & 16.0 & 28.0 \\
\midrule
\textbf{Average} & 48.2 & 47.6 & 47.8 & 43.9 & \underline{48.8} & 39.0 & \textbf{55.1} \\
\bottomrule
\end{tabular}%
}
\end{table*}

\subsection{Comparisons with State-of-the-Art Methods (Q1)}
\label{subsec:comparisons}
\noindent\textbf{Results on LIBERO Benchmark.} Following \citet{kim2024openvla}, we evaluate our model over 500 episodes for each task suite and report the average success rate. As detailed in Tab.~\ref{tab:libero_eval}, our \ourmethod achieves state-of-the-art performance with an average success rate (SR) of 98.8\%. Visual-only CoT models like CoT-VLA~\citep{zhao2025cot} lack explicit linguistic logical planning, severely restricting their ability to handle the demanding \texttt{Long} task suite. Conversely, text-only CoT models such as ThinkAct~\citep{huangthinkact} and Fast-ThinkAct~\citep{huang2026fast} struggle with spatial perception, which degrades their success on the visually intensive \texttt{Spatial} task suite. Furthermore, both paradigms predominantly rely on explicit, step-by-step autoregressive decoding, making them highly vulnerable to compounding errors that inevitably drag down overall task execution. \ourmethod addresses these limitations by internalizing joint reasoning within the continuous latent space. The Linguistic CoT provides long-horizon planning without cascading decoding errors, yielding an SR of 98.2\% on the \texttt{Long} task suite, while the Visual CoT effectively extracts geometric information, setting records of 99.4\% and 99.8\% in the \texttt{Spatial} and \texttt{Object} suites, respectively.

\noindent\textbf{Results on RoboCasa GR1 Benchmark.} Following \citet{starvla2025}, we evaluate our model over 50 episodes for each task suite and report the average success rate. Tab.~\ref{tab:robocasa_eval} outlines the evaluation on the RoboCasa GR1 benchmark, where controlling a 29-DoF dexterous hand imposes high demands on precise spatial perception. Our \ourmethod achieved an average success rate of 55.1\% across 24 tasks. The improvements are particularly prominent in spatially constrained tasks, such as \texttt{CuttingboardToPan} (80.0\%) and \texttt{PlacematToPlate} (74.0\%). This confirms that implicitly distilling dense 3D spatial priors equips the policy with the geometric grounding required for high-dimensional dexterous manipulation.

\begin{figure*}[!t]
\centering
\includegraphics[width=1.0\linewidth]{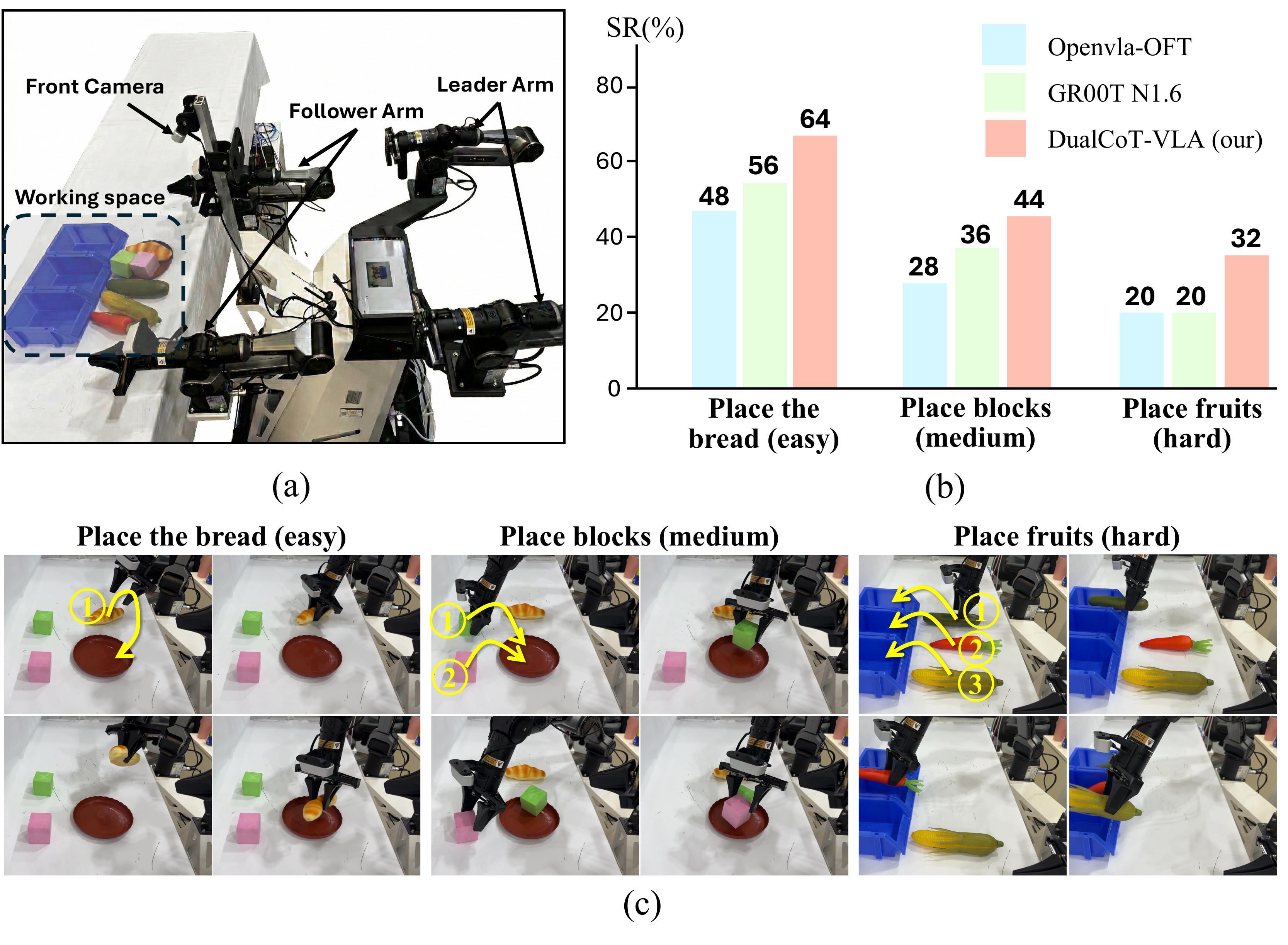}
\caption{\textbf{Real-world Experiments.} (a) Our robotic arm platform setup. (b) Success rates (SR\%) of our proposed \ourmethod compared to baseline methods (OpenVLA-OFT~\citep{KimM1-RSS-25} and GR00T-N1.6~\citep{bjorck2025gr00t}) across three real-world manipulation tasks of varying complexity: placing bread (easy), placing blocks (medium), and placing fruits (hard). (c) Qualitative execution sequences using our \ourmethod across the three manipulation tasks with increasing levels of difficulty.}
\label{fig:real_world}
\end{figure*}

\noindent\textbf{Results on Real-world.} We deploy our \ourmethod on a physical robotic arm platform (see Fig.~\ref{fig:real_world}(a)). Real-world environments are inherently unstructured, presenting challenges such as varied lighting, arbitrary object placements, and complex physical interactions. To evaluate the robustness of our method, we conduct 25 evaluation trials for each task and report the success rate (see Fig.~\ref{fig:real_world}(b)).

During physical deployment, \ourmethod demonstrates effective sim-to-real transferability across long-horizon tabletop tasks. By processing reasoning within the continuous latent space, the system sustains a high-frequency control loop, avoiding the high inference delays. The implicit Visual CoT provides essential spatial perception to minimize grasp failures, while the Linguistic CoT facilitates the sequential execution of multi-step plans. Consequently, \ourmethod yields higher success rates compared to baselines such as OpenVLA-OFT and GR00T-N1.6 across real-world tasks of varying complexity, confirming its viability as a responsive and effective control paradigm for physical robotics.

\subsection{Qualitative Analysis of Latent Reasoning (Q2)}
\label{subsec:qualitative_analysis}
To verify whether \ourmethod acquires and internalizes these reasoning capabilities, we conduct a qualitative visual assessment (see Fig.~\ref{fig:exp_vis}). The qualitative result demonstrates that our \ourmethod successfully performs multimodal reasoning  within the latent space.

To verify that our \ourmethod internalizes these reasoning capabilities, we conduct a qualitative visual assessment (Fig.~\ref{fig:exp_vis}). Specifically, we train a lightweight visual probe to map the compressed hidden states of the visual query tokens into depth maps. As shown in the middle row of Fig.~\ref{fig:exp_vis}, the probe successfully reconstructs these maps, capturing the geometric information of the visual observation. This confirms that the visual query tokens effectively extract spatial perception within the continuous latent space.

Also, we utilized the Qwen3-0.6B auxiliary decoder to explicitly translate the hidden states of the linguistic query tokens to natural language. As shown in the bottom row of Fig.~\ref{fig:exp_vis}, the decoded text matches our designated three-part CoT structure: it accurately tracks the dynamic stage of the scene, spatially localizes the relevant objects, and formulates a logical plan for the next sub-task. 

\begin{table}[!t]
\centering
\footnotesize
\caption{Detailed inference latency breakdown. The execution time of the VLM forward pass, the action head, and the total inference latency in milliseconds (ms) are reported.}
\label{tab:latency_breakdown}
\setlength{\tabcolsep}{6pt}
\begin{tabular}{lccc}
\toprule
\textbf{Metric} & Non-CoT & AR CoT & \textbf{Parallel CoT (our)} \\
\midrule
\textbf{VLM Forward} & 53.7 & 3156.0 & 58.1 \\
\textbf{Action Head} & 22.5 & 27.5 & 25.1 \\
\textbf{Total Time} & 76.2 & 3178.5 & 83.2 \\
\bottomrule
\end{tabular}
\end{table}

\begin{table}[!t]
\centering
\footnotesize
\caption{Ablation study of the impact of the Visual CoT and Linguistic CoT  across four task suites on the LIBERO benchmark~\citep{liu2023libero}. Success rates (\%) are reported.}
\label{tab:ablation}
\setlength{\tabcolsep}{5pt}
\begin{tabular}{ccccccc}
\toprule
\multicolumn{2}{c}{\textbf{CoT}} & \multicolumn{5}{c}{\textbf{LIBERO Benchmark}} \\
\cmidrule(r){1-2} \cmidrule(l){3-7}
Visual & Linguistic & \texttt{Spatial} & \texttt{Object} & \texttt{Goal} & \texttt{Long} & Avg. \\
\midrule
$\times$   & $\times$       & 97.8    & 98.8   & 97.4 & 92.0 & 96.5 \\
$\checkmark$& $\times$      & 99.4    & 99.6   & 97.4 & 95.0 & 97.9 \\
$\times$   & $\checkmark$   & 98.4    & 98.4   & 96.6 & 96.0 & 97.4 \\
$\checkmark$& $\checkmark$  & \textbf{99.4} & \textbf{99.8} & \textbf{97.8} & \textbf{98.2} & \textbf{98.8} \\
\bottomrule
\end{tabular}
\end{table}

\subsection{Analysis of Inference Speed (Q4)}
\label{subsec:inference_speed}
Inference latency is a critical limitation for autoregressive CoT VLA models in closed-loop robotic control. We present an analysis of the inference latency of our \ourmethod, both theoretically and experimentally. Theoretically, autoregressive CoT requires sequential decoding, where an N-length sequence requires $\mathcal{O}(N)$ sequential forward passes, resulting in linear scaling that is prohibitive for high-frequency control. \ourmethod structurally bypasses this by parallelizing reasoning, allowing temporal complexity to $\mathcal{O}(1)$. 

To experimentally validate this design, we conduct an inference latency analysis comparing three architectures that share an identical VLM backbone and action head: a standard Non-CoT VLA trained without reasoning, an autoregressive CoT VLA trained with sequential linguistic reasoning, and our \ourmethod. All evaluations are conducted on a single H100 GPU. As detailed in Tab.~\ref{tab:latency_breakdown}, while autoregressive CoT increases VLM forward time to 3156.0 ms, our \ourmethod increases VLM inference time by only 4.4 ms over a Non-CoT baseline (58.1 ms vs. 53.7 ms). Integrated with the DiT action head, \ourmethod achieves a rapid total latency of 83.2 ms, enabling high-frequency inference with comprehensive visual and linguistic reasoning.

\subsection{Ablation Study (Q3)}
\label{subsec:ablation}
To evaluate the individual contributions of the Visual and Linguistic CoT, we train separate model variants with distinct architectures and evaluate them on the LIBERO benchmark. As shown in Tab.~\ref{tab:ablation}, the model trained with the Visual-only CoT significantly improves performance on visually demanding tasks, reaching 99.4\% on the \texttt{Spatial} suite. The model trained with Linguistic-only CoT specifically enhances the performance on the \texttt{Long} suite, increasing the success rate from 92.0\% to 96.0\%. Our complete \ourmethod achieves the highest performance across all categories with an average success rate of 98.8\%. Overall, these experiments demonstrate that the two reasoning streams are highly complementary: the linguistic CoT handles long-horizon logical planning, while the visual CoT provides the geometric preception.

%% file: sections/discussions.tex

%% file: sections/conclusion.tex
\section{Conclusion}
\label{sec:conclusion}

We proposed a parallelized Visual-Linguistic Chain of Thought for VLA models as a dual-stream reasoning paradigm for robotic manipulation, instantiated in \ourmethod. By shifting the reasoning process from explicit autoregressive inference into the continuous latent space, our model seamlessly integrates low-level 3D spatial perception with high-level logical task planning. To achieve this, we introduced a parallel implicit CoT mechanism driven by learnable query tokens, enabling VLA models to acquire geometric perception and logical planning. our \ourmethod addresses the fundamental limitations of existing CoT-based VLAs: it overcomes the isolation of single-modal CoT to achieve a comprehensive understanding of both visual and linguistic modalities, and completely bypasses the high latency and compounding errors of autoregressive inference via single-step reasoning in the latent space. Extensive experiments across simulation benchmarks and real-world platforms validate the effectiveness of our approach, confirming that comprehensive, latent-space multimodal reasoning serves as a highly efficient and powerful inductive bias for generalist robotic control.